\documentclass[11pt,a4paper]{article}
\usepackage[hyperref]{emnlp2018}
\usepackage{times}
\usepackage{latexsym}

\usepackage{url}

\usepackage{xcolor}
\usepackage{soul}
\usepackage[utf8]{inputenc}
\usepackage[small]{caption}
\usepackage{amsmath}
\usepackage{CJK}
\usepackage{amsfonts}
\usepackage{multirow}
\usepackage{graphicx}

\usepackage{array}
\usepackage{enumitem}

\usepackage{algorithm}
\usepackage{algorithmic}
\usepackage{appendix}
\usepackage{float}
\usepackage{multicol}

\aclfinalcopy % Uncomment this line for the final submission

\title{Chinese Pinyin Aided IME, Input What You Have Not Keystroked Yet}

\author{
	Yafang Huang$^{1,2}$,
	Hai Zhao$^{1,2,*}$,
	\\
	$^1$Department of Computer Science and Engineering,\\
	Shanghai Jiao Tong University, Shanghai, 200240, China\\
	$^2$Key Laboratory of Shanghai Education Commission for Intelligent Interaction\\
	and Cognitive Engineering, Shanghai Jiao Tong University, Shanghai, 200240, China\\
	huangyafang@sjtu.edu.cn,
	zhaohai@cs.sjtu.edu.cn
	\thanks{$\ $ Corresponding author. This paper was partially supported by National Key Research and Development Program of China (No. 2017YFB0304100), National Natural Science Foundation of China (No. 61672343 and No. 61733011), Key Project of National Society Science Foundation of China (No. 15-ZDA041), The Art and Science Interdisciplinary Funds of Shanghai Jiao Tong University (No. 14JCRZ04), and the joint research project with Youtu Lab of Tencent.}
}

\date{}

\begin{document}
\maketitle
\begin{abstract}
  Chinese pinyin input method engine (IME) converts pinyin into character so that Chinese characters can be conveniently inputted into computer through common keyboard. IMEs work relying on its core component, pinyin-to-character conversion (P2C). Usually Chinese IMEs simply predict a list of character sequences for user choice only according to user pinyin input at each turn. However, Chinese inputting is a multi-turn online procedure, which can be supposed to be exploited for further user experience promoting. This paper thus for the first time introduces a sequence-to-sequence model with gated-attention mechanism for the core task in IMEs. The proposed neural P2C model is learned by encoding previous input utterance as extra context to enable our IME capable of predicting character sequence with incomplete pinyin input. Our model is evaluated in different benchmark datasets showing great user experience improvement compared to traditional models, which demonstrates the first engineering practice of building Chinese aided IME.
\end{abstract}

\section{Introduction}

\begin{figure*}[!h]
  \centering
  \includegraphics[width=1.0\textwidth]{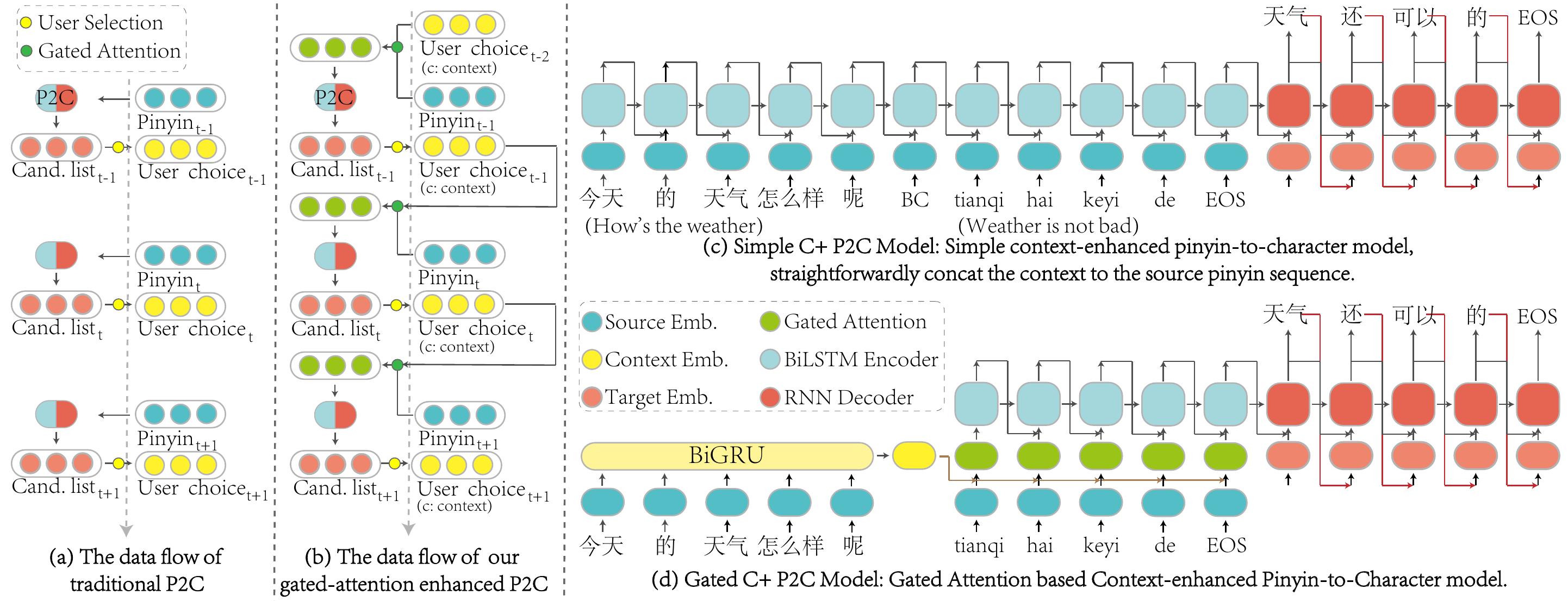}
  \caption{Architecture of the proposed model.}
  \label{fig:framework}
\end{figure*}

Pinyin is the official romanization representation for Chinese and the P2C converting the inputted pinyin sequence to Chinese character sequence is the most basic module of all pinyin based IMEs.

Most of the previous research \cite{chen:2003, zhang:2006rule, lin2008a, Chen2000A, Jiang2007Pinyin, cai:2017a} for IME focused on the matching correspondence between pinyin syllables and Chinese characters. %\cite{chen:2003} introduced a conditional maximum entropy model for grapheme-to-phoneme conversion. \cite{zhang:2006rule} presented a rule-based error correction approach to improve preferable conversion rate. \cite{lin2008a} presented a statistical model with supporting context to improve Chinese input. \cite{Chen2000A} introduced a trigram statistical language model and word segmentation for P2C. \cite{Jiang2007Pinyin} put forward a PTC framework based on support vector machine. 
\cite{huang:2018acl, yangandzhao:2013, jiaandzhao:2014, Shenyuan:2015}  regarded the P2C as a translation between two languages and solved it in statistical or neural machine translation framework. The fundamental difference between  \cite{Shenyuan:2015} work and ours is that our work is a fully end-to-end neural IME model with extra attention enhancement, while the former still works on traditional IME only with converted neural network language model enhancement. \cite{Zhang2017Tracing} introduced an online algorithm to construct appropriate dictionary for P2C. All the above mentioned work, however, still rely on a complete input pattern, and IME users have to input very long pinyin sequence to guarantee the accuracy of P2C module as longer pinyin sequence may receive less decoding ambiguity.

The Chinese IME is supposed to let user input Chinese characters with least inputting cost, i.e., keystroking, which indicates extra content predication from incomplete inputting will be extremely welcomed by all IME users. \cite{Huang2015A} partially realized such an extra predication using a maximum suffix matching postprocessing in vocabulary after SMT based P2C to predict longer words than the inputted pinyin.

To facilitate the most convenience for such an IME, in terms of a sequence to sequence model as neural machine translation (NMT) between pinyin sequence and character sequence, we propose a P2C model with the entire previous inputted utterance confirmed by IME users being used as a part of the source input. When learning the type of the previous utterance varies from the previous sentence in the same article to the previous turn of utterance in a conversation, the resulting IME will make amazing predication far more than what the pinyin IME users actually input.

In this paper, we adopt the attention-based NMT framework in \cite{Luong2015Effective} for the P2C task. In contrast to related work that simply extended the source side with different sized context window to improve of translation quality \cite{tiedemann:2017}, we add the entire input utterance according to IME user choice at previous time (referred to the \emph{context} hereafter). Hence the resulting IME may effectively improve P2C quality with the help of extra information offered by context and support incomplete pinyin input but predict complete, extra, and corrected character output. The evaluation and analysis will be performed on two Chinese datasets, include a Chinese open domain conversations dataset for verifying the effectiveness of the proposed method.

\section{Model}

As illustrated in Figure \textbf{\ref{fig:framework}}, the core of our P2C is based the attention-based neural machine translation model that converts at word level. Still, we formulize P2C as a translation between pinyin and character sequences as shown in a traditional model in Figure \ref{fig:framework}(a). However, there comes a key difference from any previous work that our source language side includes two types of inputs, the current source pinyin sequence (noted as $P$) as usual, and the extended context, i.e., target character sequence inputted by IME user last time (noted as $C$). As IME works dynamically, every time IME makes a predication according to a source pinyin input, user has to indicate the 'right answer' to output target character sequence for P2C model learning. This online work mode of IMEs can be fully exploited by our model whose work flow is shown in Figure \ref{fig:framework}(b).

As introduced a hybrid source side input, our model has to handle document-wide translation by considering discourse relationship between two consecutive sentences. The most straightforward modeling is to simply concatenate two types of source inputs with a special token 'BC' as separator. Such a model is in Figure \ref{fig:framework}(c). However, the significant drawback of the model is that there are a slew of unnecessary words in the extended context (previous utterance) playing a noisy role in the source side representation.

To alleviate the noise issue introduced by the extra part in the source side, inspired by the work of \cite{Dhingra2016Gated, pang:2016, zhang:20181, zhang:20182, zhang:20183, cai:2017acl}, our model adopts a gated-attention (GA) mechanism that performs multiple hops over the pinyin with the extended context as shown in Figure \ref{fig:framework}(d). In order to ensure the correlation between each other, we build a parallel bilingual training corpus and use it to train the pinyin embeddings and the Chinese embeddings at once. We use two Bidirectional gated recurrent unit (BiGRU) \cite{Cho2014Learning} to get contextual representations of the source pinyin and context respectively, $H_{p} = {\rm BiGRU}(P), H_{c} = {\rm BiGRU}(C)$,  where the representation of each word is formed by concatenating the forward and backward hidden states.

For each pinyin $p_i$ in $H_{p}$, the GA module forms a word-specific representation of the context $c_i \in H_{c}$ using soft attention, and then adopts element-wise product to multiply the context representation with the pinyin representation. $\alpha_i = softmax(H_{c}^\mathrm{T} p_i),  \beta_i = C\alpha_i, x_{i} = p_i \odot \beta_i$, where $\odot$ is multiplication operator.

The pinyin representation $\tilde{H}_{p} = {x_1, x_2, . . . , x_k}$ is augmented by context representation and then sent into the encoder-decoder framework. The encoder is a bi-directional long short-term memory (LSTM) network \cite{Hochreiter1997Long}. The vectorized inputs are fed to forward and backward LSTMs to obtain the internal representation of two directions. The output for each input is the concatenation of the two vectors from both directions. Our decoder based on the global attentional models proposed by \cite{Luong2015Effective} to  consider the hidden states of the encoder when deriving the context vector. The probability is conditioned on a distinct context vector for each target word. The context vector is computed as a weighted sum of previous hidden states. The probability of each candidate word as being the recommended one is predicted using a softmax layer over the inner-product between source embeddings and candidate target characters.

This work belongs to one of the first line which fully introduces end-to-end deep learning solution to the IME implementation following a series of our previous work \cite{zhu:2018lingke, wu:2018finding, zhang:2018one, bai:2018deep, cai:2018afull, he:2018acl, qin:2018acl, li:2018seq2seq, Jia2013graph}.

\section{Experiment}

\subsection{Definition of Incomplete Input for IME}
The completeness in IME is actually uneasily well-defined as it is a relative concept for inputting procedure. Note that not until user types the return key \emph{enter}, user will not (usually) really make the input choice. Meanwhile, even though the entire/complete input can be strictly defined by the time when user types \emph{enter}, user still can make decision at any time and such incompleteness cannot be well evaluated by all the current IME metrics. As the incomplete from is hard to simulate and it is diverse in types, we have to partially evaluate it in the following two ways\footnote{Our code is at https://github.com/YvonneHuang/gaIME},

\paragraph{The incomplete pinyin as abbreviation pinyin}

To compare with previous work directly, we followed \cite{Huang2015A} and focused on the abbreviated pinyin (the consonant letter only) to perform evaluation (i.e., tian$\_$qi to t$\_$q).

\paragraph{Take incomplete user input as the incomplete}

As IME works as an interactive system, it will always give prediction only if users keep typing. If user's input does not end with typing \emph{enter}, we can regard the current input pinyin sequence is an incomplete one. 

\subsection{Datasets and Settings}
\begin{table}[!h]
  \centering 
  {
    \begin{tabular}{l|cc|cc}
      \hline
      & \multicolumn{2}{c|}{PD} & \multicolumn{2}{c}{DC} \\\cline{2-5}
      & Train & Test & Train & Test \\ \cline{2-5}
      \hline
      $\#$ Sentence & 5.0M & 2.0K & 1.0M & 2.0K \\ \cline{2-5}
      L $<$ 10 $\%$ & 88.7 & 89.5 & 43.0 & 54.0 \\ \cline{2-5}
      L $<$ 50 $\%$ & 11.3 & 10.5 & 47.0 & 42.0 \\ \cline{2-5}
      L $>$ 50 $\%$ & 0.0 & 0.0 & 4.0 & 2.0 \\ \cline{2-5}
      Relativity $\%$ & 18.0 & 21.1 & 65.8 & 53.4 \\ \cline{2-5}
      \hline
    \end{tabular}
  }
  \caption{\label{tab:dataset} Data statistics for the sentence number and sentence length (L) of two corpora.}
\end{table}

\begin{table*}[!h]
	\centering 
	{
		\begin{tabular}{r|cc|cc}
			\hline
			\hline
			& \multicolumn{2}{c|}{DC} & \multicolumn{2}{c}{PD} \\\cline{2-5}
			& Top-1 & KySS & Top-1 & KySS \\\cline{5-5}
			\hline
			CoCat$\star$ & 49.72 & 0.7393 & 48.78 & 0.7868 \\
			Basic P2C$\star$ & 52.31 & 0.7305 & 47.95 & 0.7879 \\
			Simple C+ P2C$\star$ & 23.83 & 0.5431 & 43.95 & 0.7495 \\
			Gated C+ P2C$\star$ & 53.76 & 0.8896 & 48.31 & 0.7931\\
			\hline
			CoCat & 59.15 & 0.7651 & 61.42 & 0.7933 \\
			Basic P2C & 71.31 & 0.8845 & 70.5 & 0.8301\\
			Simple C+ P2C & 61.28 & 0.7963 & 60.87 & 0.7883 \\
			Gated C+ P2C & \textbf{73.89} & \textbf{0.8935} & \textbf{70.98} & \textbf{0.8407} \\
			\hline
			\hline
		\end{tabular}
	}
	\caption{\label{tab:result1} The effect of P2C modules with Different Input Forms. $\star$ means that the input is incomplete.}
\end{table*}

Our model is evaluated on two datasets, namely the People's Daily (PD) corpus and Douban conversation (DC) corpus. The former is extracted from the People's Daily from 1992 to 1998 that has word segmentation annotations by Peking University. The DC corpus is created by \cite{wu:2017sequenctial} from Chinese open domain conversations. One sentence of the DC corpus contains one complete utterance in a continuous dialogue situation. The statistics of two datasets is shown in Table \ref{tab:dataset}. The relativity refers to total proportion of sentences that associate with contextual history at word level. For example, there are 65.8$\%$ of sentences of DC corpus have words appearing in the context. With character text available, the needed parallel corpus between pinyin and character texts is automatically created following the approach proposed by \cite{yangandzhao:2013}.
% In addition, converting word to pinyin consists of initial consonants can be applied to the modeling of incomplete input in our experiments.

Our model was implemented using the PyTorch\footnote{https://github.com/pytorch/pytorch} library, here is the hyperparameters we used: (a) the RNNs used are deep LSTM models, 3 layers, 500 cells, (c) 13 epoch training with plain SGD and a simple learning rate schedule - start with a learning rate of 1.0; after 9 epochs, halve the learning rate every epoch, (d) mini-batches are of size 64 and shuffled, (e) dropout is 0.3. Word embeddings are pre-trained by word2vec \cite{mikolov:2013} toolkit on the adopted corpus and unseen words are assigned unique random vectors. (f) the gated attention layers size is 3, the hidden units number of BiGRU is 100.

Two metrics are used: Maximum Input Unit (MIU) accuracy \cite{Zhang2017Tracing} and KeyStroke Score (KySS) \cite{Jia2013Kyss}. The former measures the conversion accuracy of MIU, whose definition is the longest uninterrupted Chinese character sequence during inputting. As the P2C conversion aims to output a rank list of the corresponding character sequences candidates, the Top-$K$ MIU accuracy means the possibility of hitting the target in the first $K$ predicted items. The KySS quantifies user experience by using keystroke count. For an ideal IME with complete input, we have KySS $= 1$. An IME with higher KySS is supposed to perform better.

\subsection{Model Definition}

We considered the following baselines: (a) Google IME: the only commercial Chinese IME providing a debuggable API in the market now; (b) OMWA: online model for word acquisition proposed by \cite{Zhang2017Tracing}; (c) CoCat: an SMT based input method proposed by \cite{Huang2015A} that supports incomplete pinyin inputs.

Three models with incomplete or complete inputs will be evaluated: (a) Basic P2C, the basic P2C based on attention-NMT model; (b) Basic C2C, the basic C to C model based on Seq2Seq model; (b) Simple C+ P2C, the simple concatenated P2C conversion model that concatenate context to pinyin representation; (c) Gated C+ P2C, our gated attention based context-enhanced pinyin-to-character model. Pinyin in model * has been actually set to abbreviation form when we say it goes to \cite{Huang2015A} incomplete definition.

\subsection{Result and Analysis}

\begin{figure}
	\centering
	\includegraphics[width=0.45\textwidth]{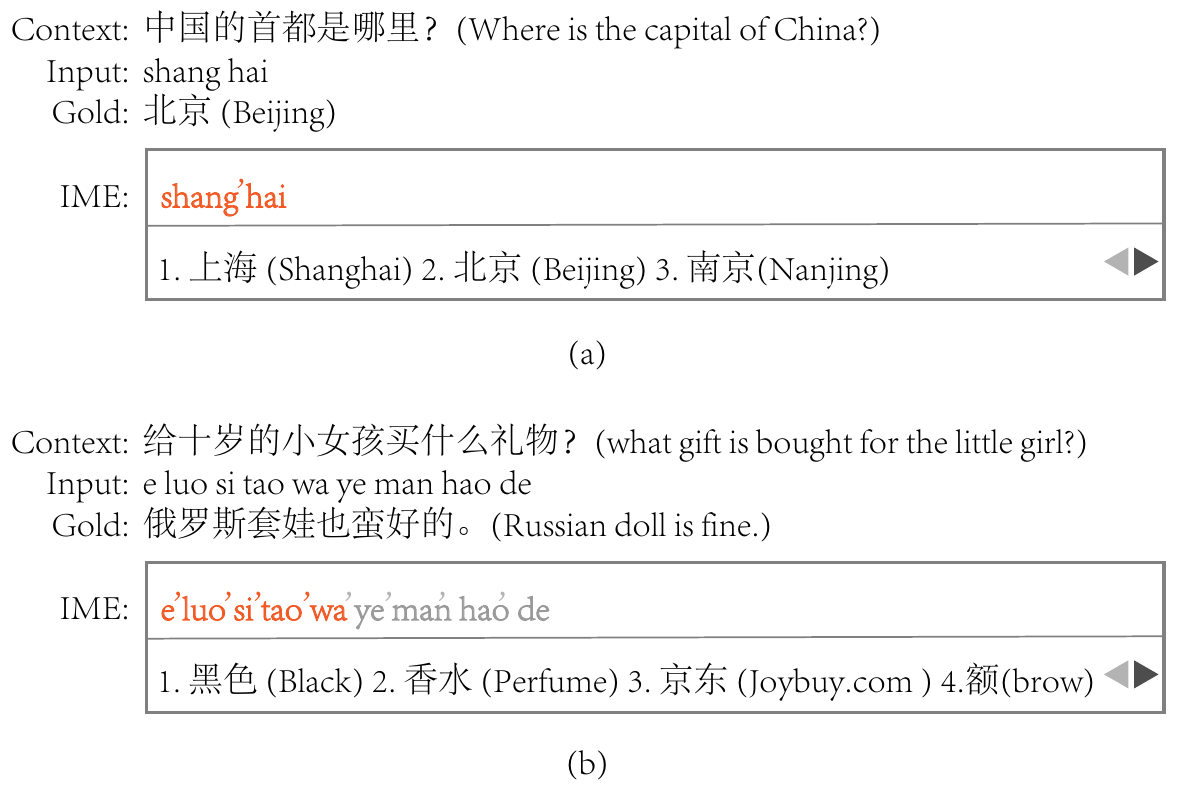}
	\caption{Examples of the candidates list given by the proposed IMEs.}
	\label{fig:interface}
\end{figure}

\begin{figure}[!h]
	\centering
	\includegraphics[width=0.47\textwidth]{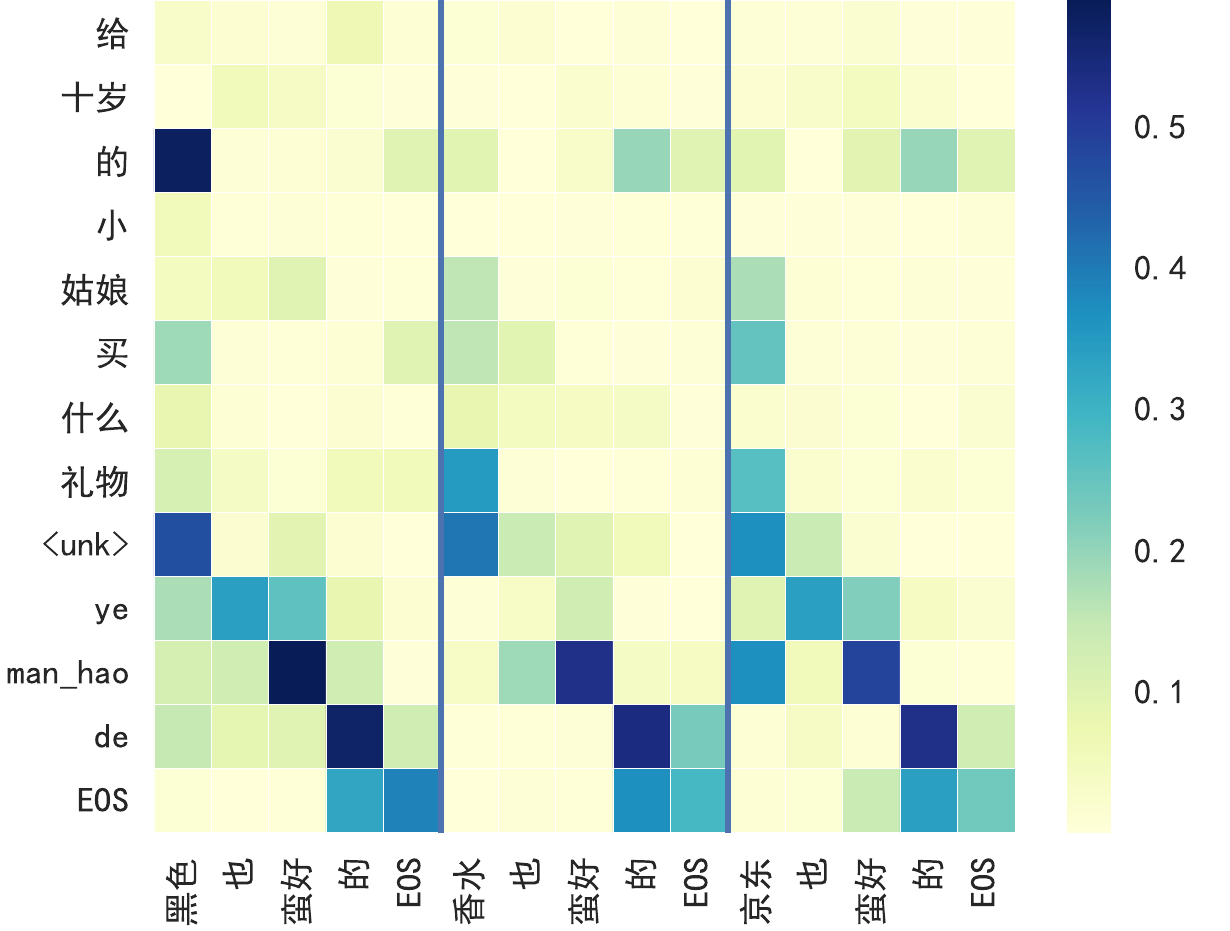}
	\caption{Attention visualization. Deeper color mean larger value.}
	\label{fig:attention}
\end{figure}

\paragraph{Effect of Gated Attention Mechanism}

Table \ref{tab:result3} shows the Effect of gated attention mechanism. We compared models with Gated C+ P2C and Simple C+ P2C. The MIU accuracy of the P2C model has over 10$\%$ improvement when changing the operate pattern of the extra information proves the effect of GA mechanism. The Gated C+ P2C achieves the best in DC corpus, suggesting that the gated-attention works extremely well for handling long and diverse context.

\paragraph{Effect of P2C modules with Different Input Forms}

\begin{table*}[!h]
	\centering 
	{
		\begin{tabular}{r|cccc|cccc}
			\hline
			\hline
			& \multicolumn{4}{c|}{DC} & \multicolumn{4}{c}{PD}\\\cline{2-9}
			& Top-1 & Top-5 & Top-10 & KySS & Top-1 & Top-5 & Top-10 & KySS \\\cline{2-9}
			\hline
			Google IME & 62.13 & 72.17 & 74.72 & 0.6731 & 70.93 & 80.32& \textbf{82.23} & 0.7535 \\
			CoCat & 59.15 & 71.85 & 76.78 & 0.7651 & 61.42 & 73.08 & 78.33 & 0.7933 \\
			OMWA & 57.14 & 72.32 & 80.21 & 0.7389 & 64.42 & 72.91 & 77.93 & 0.7115\\
			\hline
			basic P2C & 71.31 & 89.12 & 90.17 & 0.8845 & 70.5 & 79.8 & 80.1 & 0.8301\\
			Simple C+ P2C & 61.28 & 71.88 & 73.74 & 0.7963 & 60.87 & 71.23 & 75.33& 0.7883 \\
			Gated C+ P2C & \textbf{73.89} & \textbf{90.14} & \textbf{91.22} & \textbf{0.8935} & \textbf{70.98} & \textbf{80.79} & 81.37 & \textbf{0.8407} \\ \cline{2-9}
			\hline
			\hline
		\end{tabular}
	}
	\caption{\label{tab:result3} Comparison with previous state-of-the-art models.}
\end{table*}

Table \ref{tab:result1} shows the evaluation results of P2C modules with different input forms. It should not surprise that straightforward concatenation strategy for source inputs performs poorly when the input pinyin is incomplete in DC corpus, due to obvious noise in too long context. The relatively small gap between the results of CoCat$\star$ and CoCat indicate that statistical learning model may be helpful in obtaining some useful patterns from limited input. When the input statement contains adequacy information, the MIU accuracy of Gated C+ P2C system achieves more than 20$\%$ improvement in both corpora. However, we find that the KySS scores are much more close even with different pinyin integrity, which indicates that user experience in terms of KySS are more hard improved.

\paragraph{Instance Analysis}

We input a dialogue in wonder to how much of the contextual information is used when P2C module find the input pinyin is unknown. Figure \ref{fig:interface} demonstrates the effect of the gated attention mechanism on candidates offering and unknown word replacement. As shown in Figure \ref{fig:interface}(a), we find that our IME suggests a more suitable candidates to the user when user is obviously not consistent with what the model has learned previously, which shows our model exceeds the Simple C+ P2C learning for maximally matching the inputted pinyin, but become capable of effectively resisting user pinyin input noise, and turns to learn potential language knowledge in previous input history\footnote{Note as we evaluate our model only on two available corpora, but not the real world case from true user inputting history, which makes the instance situation limit to the domain feature of the given corpora.}. 

As the ability predict user input from incomplete pinyin cannot be covered by any current IME performance metrics, thus the reported results yielded by our model actually underestimate our model performance to some extent. We illustrate the empirical discoveries of Figure \ref{fig:interface}(b) to demonstrate the extra effect of our P2C system on such situation, which indicates that the gated-attention pattern has taken great advantage of contextual information when given an unknown word.  Or, namely, our model enables the incomplete input prediction though has to let it outside the current IME performance measurement.
%The candidates list in Figure \ref{fig:interface}(b) indicates that the gated-attention pattern has taken great advantage of contextual information when given an unknown word. 
We display the attention visualization of Figure \ref{fig:interface}(b) in Figure \ref{fig:attention} for better reference to explain the effect extended context plays on the generation of target characters.

\paragraph{Main Result}

Our model is compared to other models in Table \ref{tab:result3}. So far, \cite{Huang2015A} and \cite{Zhang2017Tracing} reported the state-of-the-art results among statistical models. We list the top-5 accuracy contrast to all baselines with top-10 results, and the comparison indicates the noticeable advancement of our P2C model. To our surprise, the top-5 result  on PD of our best Gated C+ P2C system approaches the top-10 accuracy of Google IME. On DC corpus, our Gated C+ P2C model with the best setting achieves 90.14$\%$ accuracy, surpassing all the baselines. The comparison shows our gated-attention system outperforms all state-of-the-art baselines with better user experience.

\section{Conclusion}

For the first time, this work makes an attempt to introduce additional context in neural pinyin-to-character converter for pinyin-based Chinese IME as to our best knowledge. We propose a gated-attention enhanced model for digging significant context information to improve conversion quality. More importantly, the resulting IME supports incomplete user pinyin input but returns complete, extra and even corrected character outputs, which brings about a story-telling mode change for all existing IMEs. 
%The evaluation on standard linguistic corpus and open domain conversion corpus shows the proposed methods indeed greatly improve user experience in terms of diverse metrics compared to commercial IME and state-of-the-art traditional models.
%\addtolength{\textheight}{0cm}

\bibliography{emnlp2018}

%\addtolength{\textheight}{-17cm}

\twocolumn[{
%\addtolength{\textheight}{-17cm}
\begingroup
\renewcommand{\section}[2]{}

\endgroup

\bibliographystyle{acl_natbib_nourl}

%\twocolumn[{
\appendix
\appendixpage
\section{Architecture of the Evaluation Models}
	\begin{figure}[H]
		\centering
		\includegraphics[height=0.60\textheight]{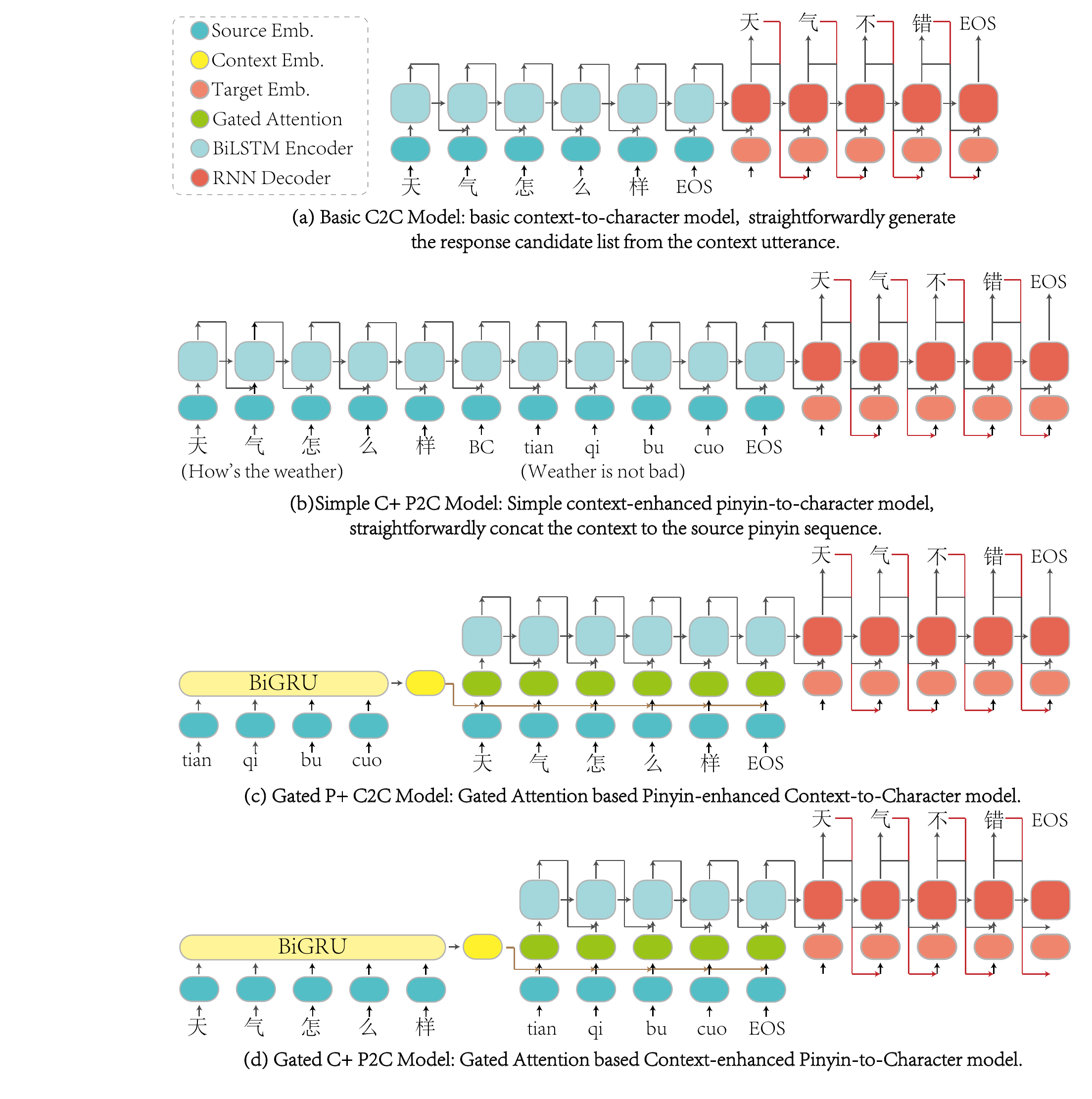}
		%\caption{Complete architecture of the evaluation model.}
		\label{fig:fwapp}
	\end{figure}
}]

\end{document}